\pdfoutput=1

\documentclass[11pt]{article}

\usepackage[]{EMNLP2022}

\usepackage{times}
\usepackage{latexsym}
\usepackage{multirow}
\usepackage{graphicx}
\usepackage{amssymb}
\usepackage{booktabs}
\usepackage{float}
\usepackage{xcolor, colortbl}

\usepackage[T1]{fontenc}

\usepackage[utf8]{inputenc}

\usepackage{microtype}

\usepackage{inconsolata}

\usepackage{adjustbox}
\title{Controllable Lexical Simplification for English}

\author{Kim Cheng Sheang, Daniel Ferrés, Horacio Saggion \\
        LaSTUS Lab, TALN Group, Universitat Pompeu Fabra \\
        C/Roc Boronat 138, Barcelona, 08018, Spain \\
        \{kimcheng.sheang, daniel.ferres, horacio.saggion\}@upf.edu }

\begin{document}
\maketitle
\begin{abstract}

Fine-tuning Transformer-based approaches have recently shown exciting results on sentence simplification task. However, so far, no research has applied similar approaches to the Lexical Simplification (LS) task. In this paper, we present ConLS, a Controllable Lexical Simplification system fine-tuned with T5 (a Transformer-based model pre-trained with a BERT-style approach and several other tasks). The evaluation results on three datasets (LexMTurk, BenchLS, and NNSeval) have shown that our model performs comparable to LSBert (the current state-of-the-art) and even outperforms it in some cases. We also conducted a detailed comparison on the effectiveness of control tokens to give a clear view of how each token contributes to the model.
\end{abstract}

\section{Introduction}
Lexical  Simplification (LS) is a Natural Language Processing task that modifies texts by substituting difficult words with easier words (or phrases) while keeping the original information and meaning \cite{shardlowsurvey2014}. Table \ref{table:an_LS_example} shows an example of a lexical simplification. On the other hand, Syntactic Simplification (SS) is a similar task that reduces the syntactic complexity of a text. Both LS and SS tasks can be seen as sub-tasks of the broader task of Automatic Text Simplification \cite{Saggion'2017}, which reduces both the lexical and syntactic complexity of texts. Lexical Simplification systems \cite{paetzoldsurvey2017} usually have components for 1) identification of complex words; 2) generation of substitution words; 3) selection of the substitutes that can fit in the context; 4) ranking substitutes by their simplicity; and 5) morphological and contextual adaptation (if necessary). The systems evaluated in this paper do not perform complex word identification. We use datasets that already had a complex word tagged for each instance. Moreover, we do not address the morphological and context adaptation task because neural-based language models usually return a correct inflected candidate. 

 \begin{table}[ht]
        \centering
        \def\arraystretch{1.1}
        \begin{tabular}{p{0.94\linewidth}}
        \specialrule{1.1pt}{1pt}{1pt}
        \textbf{Complex Sentence}: \\
        The Hush Sound is currently on \textbf{hiatus}.
         \\ \hline
         \textbf{Simplified Sentence}: \\ 
        The Hush Sound is currently on \textbf{break}.\\
        \specialrule{1.1pt}{1pt}{1pt}
        \end{tabular}
        \caption{A lexical simplification example taken from the LexMTurk dataset \citep{horn-etal-2014-learning} with the complex word and the substitute word in bold.}
        \label{table:an_LS_example}
    \end{table}
    
The contributions of this paper are:

\begin{itemize}
    \item To the best of our knowledge, we are the first to introduce a controllable mechanism for LS and to fine-tune a Transformer-based model for LS. \footnote{The code and data are available at \url{https://github.com/KimChengSHEANG/ConLS}}
    \item We have conducted an extensive evaluation of several metrics. This allows us to better understand the system when applied to real-world scenarios.

\end{itemize}

The rest of the paper is organized as follows: in Section 2, we describe related work on Lexical Simplification focusing on neural-based systems. 
Section 3 presents the ConLS approach. 
Section 4 describes the evaluation metrics and presents the experimental results.
Section 5 discusses the results of the experiments, while Section 6 concludes the paper and presents future work.

\section{Related Work}

Early Lexical Simplification approaches with unsupervised models used:  Latent Words Language Models \cite{Bel:Moe:10}, Wikipedia-based models/rules \cite{Biran:11,Yatskar10,horn-etal-2014-learning} and distributional lexical semantics \cite{glavas-stajner-2015}. \cite{paetzold-specia-2017-lexical} started the use of neural networks for the task combined with a retrofitted context-aware word embedding model.

\cite{qiang2020BERTLS,QiangLSBertACM} presented LSBert, a Lexical Simplification system that uses a pre-trained BERT \cite{devlin-etal-2019-bert} model for English to generate 
 substitution candidates.  LSBert has two main phases: 1) Substitution Generation with the BERT Masked Language Model, and 2) Substitution Filtering and Ranking with several features: BERT prediction order, a BERT language model, PPDB database, corpus-based word frequency, and FastText similarity.

\citet{martin-etal-2020-controllable} presented ACCESS a controllable Text Simplification system based on Sequence-to-Sequence models. This system allows explicit control of simplification conditions such as length, amount of paraphrasing, lexical complexity, and syntactic complexity.
ACCESS achieved SOTA results in Text Simplification benchmarks on the WikiLarge test set. Later on, \citet{martin2021muss} introduced MUSS (an extended version) by fine-tuning BART \cite{bart2019} with ACCESS, and the results were improved. In addition, \citet{sheang-saggion-2021-controllable} took a similar approach, adding another control token (number of words) and fine-tuning it with T5 \cite{2020t5}.


\section{System Description}
    Following recent works of \citet{martin-etal-2020-controllable}, \citet{martin2021muss}, \citet{sheang-saggion-2021-controllable}, and \citet{stajner_sheang_saggion_2022}, we are inspired to apply a similar approach in lexical simplification task. Specifically, our model is based on \citet{sheang-saggion-2021-controllable}, a model originally developed for sentence simplification\footnote{\url{https://github.com/KimChengSHEANG/TS_T5}}. 
    We propose a controllable mechanism for LS because we believe that the embedded token values extracted from training data could give additional information to the model about the relations between the source and the target word; so that at inference, we could define different token values that fulfill our objectives, which in this case is to find the best candidates. In the following paragraphs, we describe all the details about each token and the reason why they are chosen.
    
        \paragraph{Word Length (WL)} is the character length ratio between the complex word and the target word. It is the number of characters of the target word divided by the number of characters of the complex word. Based on our analysis of the training dataset (TSAR-EN), 65.71\% of the time complex word is longer than the best candidate, 21.30\% the complex word is shorter than the best candidate, and 12.99\% both are the same length. 
        
        \paragraph{Word Rank (WR)} is the inverse frequency of the target word divided by that of the complex word. The inverse frequency order is extracted from the FastText pre-trained model. Based on our analysis of the TSAR-EN dataset, 85.45\% of the time, the complex word has a lower frequency than the best candidate. Therefore, this token is a good indicator to help guide the model to predict simpler candidates.
        
        \paragraph{Candidate Ranking (CR)} is the ranking order extracted from the training data. The values are given to candidates by the ranking order. E.g., the best-ranking candidate is given the value 1.00, the second 0.75, the third 0.50, the fourth 0.25, and starting from the fifth, it is given 0.00. We used only five different values to avoid overloading the model, as the training data is relatively small. In addition, the rationale behind using these values is that we want the model to learn candidates ranking from data through the training process rather than injecting additional information or doing post-processing.

\section{Experiments}
    In our experiments, we compare our model with the current state-of-the-art model LSBert \cite{qiang2020BERTLS}. We used the original LSBert configurations and resources, and we made the following changes to have a detailed comparison with our model. By default, LSBert returns only a single best candidate for each complex word, so we made the changes to return the 10 best-ranked candidates. We changed the number of BERT mask selections from 10 to 15 so that after removing duplicate candidates, we still have around 10 candidates. Moreover, we filtered out all the candidates that were equal to the complex word. Due to the fact that all the used datasets have gold annotated simpler substitutions in all instances, we could assume that returning the complex word would be incorrect.

\subsection{Datasets}
    This subsection describes all the Lexical Simplification datasets for English that we used in our experiments. We used LexMTurk \cite{horn-etal-2014-learning}, BenchLS\footnote{\url{https://doi.org/10.5281/zenodo.2552393}} \cite{Paetzold2016BenchmarkingLS}, and NNSeval \footnote{\url{https://doi.org/10.5281/zenodo.2552381}} \cite{Paetzold_Specia_2016UNS} for testing and TSAR-EN \cite{frontiers2022} dataset for training and validation. LexMTurk has 500 sentences that were obtained from Wikipedia. This dataset contains the marked complex words and their replacements suggested by 50 English-speaking annotators. The BenchLS dataset is a union of the LSeval \cite{BelderM12} and LexMTurk datasets in which spelling and inflection errors were automatically corrected.
    The NNSeval dataset is a filtered version of the BenchLS adapted to evaluate LS for non-native English speakers.
    
    \begin{table}[h]
        \centering
        \def\arraystretch{1.5}
        \begin{tabular}{p{0.94\linewidth}}
        \specialrule{1.1pt}{1pt}{1pt}
        \textbf{Sentence}
        \\ \hline 
        
        European Union foreign ministers agreed Monday to impose fresh sanctions on Syria as a U.N.-backed peace plan -- along with all other diplomatic efforts -- has yet to stop the \textbf{carnage} that mounts every day.
         \\ \hline
         \textbf{Simpler Substitutes} \\ \hline
        
        destruction:6, bloodshed:3, massacre:3, slaughter:3, carnage:2, brutality:1, butchering:1, butchery:1, damage:1, death:1, slaying:1, violence:1, war:1\\
        \specialrule{1.1pt}{1pt}{1pt}
        \end{tabular}
        \caption{An example taken from the TSAR-EN dataset \citet{Stajner&al'2012} with the target word in bold. The numbers after ':' represents the number of workers that suggested the substitution. Each instance has 25 substitutes suggested by 25 crowd-sourced workers. }
        \label{table:an_example_tsar_en}
    \end{table}

    TSAR-EN dataset has 386 instances with 25 gold-annotated substitutions. Table \ref{table:an_example_tsar_en} shows an example. The instances and their target complex words were extracted from the Complex Word Identification shared task 2018 \cite{yimametal2018}. The instances were annotated using Amazon’s Mechanical Turk by 25 annotators. A native English annotator reviewed all suggestions.

\subsection{Evaluation Metrics}
 We evaluated the systems with several metrics that could take into account the results for different numbers of K candidates (from 1 up to 10). 
The metrics used are the following:
\begin{itemize}

\item {\em Accuracy@1}: is the ratio of instances with the top-ranked candidate in the gold standard list of annotated candidates.  

\item {\em Accuracy@K@top1}: The ratio of instances where at least one of the top K predicted candidates matches the most frequently suggested synonym/s\footnote{Ties in the most repeated gold-annotated candidates are taken into account.} in the gold list of annotated candidates.

\item {\em Potential@K}: the percentage of instances for which at least one of the top K substitutes predicted is present in the set of gold annotations.

\item {\em Mean Average Precision@K (MAP@K)}: This metric evaluates the relevance and ranking of the top K predicted substitutes.

\item Precision@K: the percentage of top K generated candidates that are in the gold standard.

\item Recall@K: the percentage of gold-standard substitutions that are included in the top K generated substitutions.

\end{itemize}

\subsection{Experimental Setup}
    In this section, we describe how the data are preprocessed, the training details of the model, and finally, the generation of candidates.
    
\subsubsection{Data Preprocessing} 
For each instance, we have a sentence, a complex word, and a list of ranked candidates. We compute all the ratios and the ranking, then prepend it to the source sentence. We also use special tokens [T] and [/T] to mark the boundary of the complex word in the source sentence and the simple word in the target sentence. Moreover, these special tokens help us identify the candidates during the inference. 
Table \ref{table:training_example} shows an example of source and target sentences embedded with token values and boundary tokens.

\begin{table}[h]
    \centering
    \def\arraystretch{1.5}
    \begin{tabular}{p{0.94\linewidth}}
    \specialrule{1.1pt}{1pt}{1pt}
    \textbf{Source:} <CR\_1.00> <WL\_0.54> <WR\_0.90> The Obama administration has seen what The New York Times calls an [T]unprecedented[/T] crackdown on leaks of government secrets. \\

    \specialrule{0.9pt}{1pt}{1pt}
    
    \textbf{Target:} The Obama administration has seen what The New York Times calls an [T]unusual[/T] crackdown on leaks of government secrets. \\
    \specialrule{1.1pt}{1pt}{1pt}
    \end{tabular}
    \caption{A training example. The control token values are extracted from the complex word (unprecedented) and one substitute word (unusual). The word unusual is the best-ranked candidate suggested by annotators, so the CR value is 1.00. We used all the candidates in each instance to generate parallel sentences for training. One candidate per training example.}
    \label{table:training_example}
\end{table}

\begin{table*}[ht!]
\begin{adjustbox}{width=1\textwidth,center=\textwidth}
\begin{tabular}{lcclccccccccccc}
\specialrule{1.3pt}{1pt}{1pt}
\multirow{2}{*}{Dataset}   & \multirow{2}{*}{System}   & \multirow{2}{*}{ACC@1} &  & \multicolumn{5}{c}{ACC@k@Top1}   &  & \multicolumn{5}{c}{Potential@k}       \\
                           &                           &                        &  & @1    & @2    & @3    & @4    & @5    &  & @2    & @3    & @4    & @5    & @10   \\ \hline
\multirow{2}{*}{BenchLS}   & LSBert                    & \textbf{67.59}                  &  & \textbf{40.68} & \textbf{51.45} & 57.37 & 59.84 & 61.57 &  & \textbf{77.07} & \textbf{81.27} & 83.32 & 84.28 & 85.47 \\
                           & ConLS                     & 62.00                  &  & 37.99 & 51.34 & \textbf{59.31} & \textbf{64.90} & \textbf{68.46} &  & 74.92 & \textbf{81.27} & \textbf{84.82} & \textbf{87.08} & \textbf{90.31} \\ \hline
\multirow{2}{*}{NNSeval}   & LSBert                    & \textbf{44.76}                  &  & \textbf{28.03} & \textbf{38.49} & 43.93 & 46.86 & 49.79 &  & \textbf{59.00} & \textbf{64.85} & \textbf{67.78} & \textbf{71.55} & 74.48 \\
                           & ConLS                     & 41.00                  &  & 26.77 & 34.30 & \textbf{45.18} & \textbf{50.20} & \textbf{52.71} &  & 53.14 & 61.09 & 65.69 & 69.87 & \textbf{79.08} \\ \hline
\multirow{2}{*}{LexMTurk} & LSBert                    & \textbf{84.80}                  &  & \textbf{44.00} & 54.80 & 60.40 & 61.80 & 62.80 &  & \textbf{91.00} & 93.20 & 94.60 & 95.00 & 95.80 \\
                           & \multicolumn{1}{l}{ConLS} & 80.60                  &  & 43.80 & \textbf{56.39} & \textbf{65.40} & \textbf{71.20} & \textbf{76.60} &  & 90.00 & \textbf{95.60} & \textbf{97.40} & \textbf{98.20} & \textbf{99.60} \\
\specialrule{1.3pt}{1pt}{1pt}
\end{tabular}

\end{adjustbox}
	\caption{The results of LSBert and ConLS for the metrics: Accuracy@1, Accuracy@k@Top1, and Potential@K.}
	\label{table_results1}
\end{table*}

\begin{table*}[ht!]
\begin{adjustbox}{width=1\textwidth,center=\textwidth}
\begin{tabular}{lcccccccccccccc}
\specialrule{1.3pt}{1pt}{1pt}
\multirow{2}{*}{Dataset}   & \multirow{2}{*}{System} & \multicolumn{5}{c}{MAP@k}             &  & \multicolumn{4}{c}{Precision@k} & \multicolumn{3}{c}{Recall@k} \\
                           &                         & @2    & @3    & @4    & @5    & @10   &  & @3      & @5      & @10     &   & @3       & @5      & @10     \\ \hline
\multirow{2}{*}{BenchLS}   & LSBert                  & \textbf{52.26} & \textbf{42.29} & 34.79 & 29.25 & 15.74 &  & \textbf{46.46}   & 34.62   & 24.90    &   & \textbf{25.74}    & 29.80    & 32.41   \\
                           & ConLS                   & 49.73 & 41.37 & \textbf{35.01} & \textbf{30.54} & \textbf{18.84} &  & 46.34   & \textbf{37.11}   & \textbf{26.20}    &   & 25.59    & \textbf{32.25}   & \textbf{41.89}   \\ \hline
\multirow{2}{*}{NNSeval}   & LSBert                  & \textbf{34.93} & \textbf{27.84} & 23.18 & 19.97 & 10.73 &  & 32.84   & 26.16   & 18.78   &   & \textbf{19.55}    & 23.40    & 26.14   \\
                           & ConLS                   & 31.69 & 27.31 & \textbf{23.23} & \textbf{20.30}  & \textbf{12.53} &  & \textbf{32.91}   & \textbf{27.02}   & \textbf{19.51}   &   & 18.80     & \textbf{23.80}    & \textbf{32.08}   \\ \hline
\multirow{2}{*}{LexMTurk} & LSBert                  & \textbf{67.05} & 54.41 & 45.83 & 39.01 & 21.29 &  & 58.03   & 45.25   & 33.43   &   & 20.52    & 24.61   & 27.52   \\
                           & ConLS                   & 65.45 & \textbf{55.45} & \textbf{48.04} & \textbf{42.52} & \textbf{27.59} &  & \textbf{60.16}   & \textbf{49.89}   & \textbf{36.94}   &   & \textbf{21.32}    & \textbf{27.51}   & \textbf{37.15}  \\
\specialrule{1.3pt}{1pt}{1pt}

\end{tabular}

\end{adjustbox}
\caption{The results of LSBert and ConLS for the metrics: {\em MAP@K}, {\em Precision@K}, and {\em Recall@K}.}
	\label{table_results2}
\end{table*}

\subsubsection{Training} 
    For our experiments, we fine-tuned T5-Large on the TSAR-EN dataset. We also compared the differences of T5 models; the results are in Table \ref{table_model_size_results}. We split the dataset to 90\% for training and 10\% for validation. This 10\% validation set is also used in the token values search at the inference, as described in the following section. For the training data, we preprocessed by extracting and adding control tokens to the source sentence along with the boundary tokens to the complex word and substitute word, as shown in Table \ref{table:training_example}. We set the maximum sequence length (number of tokens) to 128, as all our datasets contain less than 128 in tokens length. We used Optuna \cite{Akiba2019OptunaAN} for hyper-parameters search. For more details about the implementation and hyperparameters, please check Appendix \ref{appendix:implementation_details}.

\subsubsection{Inference}
    First, we performed token values search on the validation set that maximizes the Accuracy@1@top1 score using Optuna \cite{Akiba2019OptunaAN}. We searched the values ranging between 0.5 and 1.25; at each iteration, we changed the value by 0.05. We searched only WL and WR, whereas for CR, we set it to 1.00 because we already knew that the best-ranking candidates were given the value of 1.00. Then we kept these values fixed for all sentences at the inference. Finally, at the inference, we set the beam search to 15 and the number of return sequences to 15 so that after filtering out some duplicate candidates, the remaining would be around 10. The ranking order of the candidates is chosen from the return orders of sequences produced by the model.

\section{Results and Discussion}

In Table \ref{table_results1} we present the results for the metrics: {\em Accuracy@1}, {\em Accuracy@k@Top1}, and {\em Potential@K}. In Table \ref{table_results2} we present the results for the metrics: {\em MAP@K}, {\em Precision@K}, and {\em Recall@K}. The results of ConLS presented here are based T5-Large.

Our experiments show that the modified LSBert had improved its {\em Accuracy@1} metric results with respect to the ones seen in the original LSBert paper \cite{QiangLSBertACM}:  {\em Accuracy@1} has improved from 79.20  to 84.80 for LexMTurk, from 61.60 to 67.59 for BenchLS, and from 43.60 to 44.76 for NNSeval. 
On the other hand, for the {\em Accuracy@1} metric the  ConLS system does not improve the results of the modified LSBert system but improves the results of the original LSBert for the LexMturk and BenchLS datasets. The results of the {\em Accuracy@k@Top1} metric show that the modified LSBert achieves better results at K=$\{$1,2$\}$ and the ConLS achieves better results at K=$\{$3,4,5$\}$ for all datasets. This indicates that with more candidates allowed (3, 4, and 5 candidates) the ConLS is able to generate more instances with candidates within the top-1(s) gold annotated substitution(s) with respect to LSBert. The results of the {\em Potential@K} metric show these facts: 1) in LexMturk and BenchLS, the ConLS is outperforming LSBert gradually and increasingly from k=3 to k=10; 2) in NNSeval, ConLS improves the potential of LSBert only at K=10. For the MAP@K metric, we show that ConLS is able to improve the results of the metric at K=$\{$4,5,10$\}$ in all the datasets with respect to the modified LSBert.
Finally, the results of the {\em Precision@K} and {\em Recall@K} metrics show the same pattern: 1) for LexMTurk, ConLS outperforms the LSBert in all K=$\{$3,5,10$\}$; 2) for BenchLS and NNSEval, ConLS outperforms the LSBert only in K=$\{$5,10$\}$.

\begin{table}[h]
\begin{tabular}{lllll}
\specialrule{1.3pt}{1pt}{1pt}
\multicolumn{1}{c}{\multirow{2}{*}{T5   Model}} & \multicolumn{1}{c}{\multirow{2}{*}{ACC@1}} & \multicolumn{3}{c}{ACC@k@Top1}                                           \\
\multicolumn{1}{c}{}                            & \multicolumn{1}{c}{}                       & \multicolumn{1}{c}{@1} & \multicolumn{1}{c}{@2} & \multicolumn{1}{c}{@3} \\ \hline
T5-Small                                        & 23.40                                      & 7.80                   & 11.80                  & 15.40                  \\
T5-Base                                         & 60.00                                      & 28.80                  & 40.40                  & 48.40                  \\
T5-Large                                        & \textbf{80.60}                                      & \textbf{43.80}                  & \textbf{56.39}                  & \textbf{65.40}          \\
\specialrule{1.3pt}{1pt}{1pt}

\end{tabular}
\caption{The results of ConLS trained all tokens using different T5 models. The models were trained with TSAR-EN and evaluated with LexMTurk.}
	\label{table_model_size_results}
\end{table}

We also conducted a comparison on the effect of different T5 models trained with TSAR-EN and evaluated with LexMTurk. Table \ref{table_model_size_results} shows that the T5-Large model performs a lot better than the T5-Base and the T5-Small models in all metrics (Accuracy@1, Accuracy@k@Top1). Therefore, we believe that the performance of our model would improve if we could go with larger model, for example, T5-3b or T5-11b. We have tried with T5-3b model, but unfortunately it was unable to fit into our GPU memory (NVidia RTX 3090) even though we had set the batch size to as small as one.

\begin{table}[h]
\begin{tabular}{lllll}
\specialrule{1.3pt}{1pt}{1pt}
\multirow{2}{*}{Tokens} & \multirow{2}{*}{ACC@1} & \multicolumn{3}{c}{ACC@k@Top1}     \\
&    & \multicolumn{1}{c}{@1} & \multicolumn{1}{c}{@2} & \multicolumn{1}{c}{@3} \\ \hline
No Tokens               & 79.20                  & 41.80                  & 55.20                  & 62.60                  \\
CR                      & 79.00                  & 41.00                  & 54.40                  & 62.60                  \\
WL                      & 79.40                  & 43.00                  & 55.20                  & 65.00                  \\
WR                      & 78.60                  & 41.20                  & 54.60                  & 63.20                  \\
CR+WL                   & 78.40                  & 41.40                  & 54.40                  & 62.40                  \\
CR+WR                   & 78.60                  & 42.80                  & 54.60                  & 62.20                  \\
WL+WR                   & 78.60                  & 41.00                  & 54.20                  & 62.20                  \\
All Tokens              & \textbf{80.60}                                      & \textbf{43.80}                  & \textbf{56.39}                  & \textbf{65.40}             \\
\specialrule{1.3pt}{1pt}{1pt}
\end{tabular}
\caption{The results of ConLS trained with different set of tokens. Each model was trained with TSAR-EN and evaluated with LexMTurk. }
	\label{table_token_sets_results}
\end{table}

To evaluate the effectiveness of the control tokens, we conducted further experiments with different set of combinations. We trained and evaluated each set of tokens using T5-Large with TSAR-EN for training and LexMTurk for evaluation. The results on Table \ref{table_token_sets_results} have shown that the model trained with no tokens performs lower than the model with all tokens in all metrics, especially for the Accuracy@1@Top1 metric, the model with all tokens perform +2 points higher. Moreover, the \textbf{all tokens} model performs better than all other models in all metrics. This indicates that each token contributes to the selection and the ranking of the candidates that leads to better performance.

\section{Conclusions and Future Work}

This paper presents ConLS, the first approach for Controllable Lexical Simplification. 
The paper also describes the evaluation of LSBert and ConLS for English with the LexMTurk, BenchLS, and NNSeval datasets for testing and the TSAR-EN dataset for training. The results of our evaluation show that the modified LSBert improves the {\em Accuracy@1} metric results with respect to the ones seen in the original LSBert paper in all three datasets. ConLS also improves it for the LexMturk and BenchLS datasets. Moreover, the ConLS system is able to achieve: 1) more potential to capture correct answers at K=$\{$3,4,5,10$\}$ for BenchLS and LexMturk and at K=10 for NNSeval with respect to LSBert, 2) with more candidates retrieved (4 or 5) is able to generate more candidates within the top-1 more frequent gold-annotated suggestions with respect to LSBert, 3) with K=$\{$5,10$\}$ candidates is able to generate (according to the gold-annotations) more correct and different candidates.

For future work, we plan to build a custom model to predict the best control token values from a given input instance. Having instance-customized control token values seems more adequate, as humans usually select the best candidate based on context.

\section*{Limitations}

We describe in this Section the limitation of our work. The most probable limiting features are:
\begin{itemize}
\item The size of training dataset: the TSAR-EN dataset has 386 instances. Obviously, training with datasets with a large number of instances would be recommended to create better models.
\item Quality of the training dataset: although during the creation of the TSAR-EN dataset, it was inspected and the unsuitable substitutions were removed and replaced with suitable ones \cite{frontiers2022}, it is possible that the dataset quality could be improved by including substitutions not reported by the annotators.
\item Quality of the testing datasets:  it is also possible that these datasets could be improved by including substitutions not reported by the annotators.
\item Successful adaptation to other languages: we could have possible difficulties in achieving similar adaptations and results in non-English languages due to the difficulties in availability of similar resources for other languages and specifically for low-resource languages.

\end{itemize}

\section*{Ethics Statement}
We have described the limitations of the proposed method in the previous Section.
All the scientific datasets and algorithms used are properly cited.

\section*{Acknowledgements}
Our work is supported from the project Context-aware Multilingual Text Simplification (ConMuTeS) PID2019-109066GB-I00/AEI/10.13039/501100011033 awarded by Ministerio de Ciencia, Innovación y Universidades (MCIU) and by Agencia Estatal de Investigación (AEI) of Spain. In addition, we would like to thank the anonymous reviewers for their constructive comments and suggestions.

\bibliography{custom}

\begin{thebibliography}{27}
\expandafter\ifx\csname natexlab\endcsname\relax\def\natexlab#1{#1}\fi

\bibitem[{Akiba et~al.(2019)Akiba, Sano, Yanase, Ohta, and
  Koyama}]{Akiba2019OptunaAN}
Takuya Akiba, Shotaro Sano, Toshihiko Yanase, T.~Ohta, and Masanori Koyama.
  2019.
\newblock Optuna: A next-generation hyperparameter optimization framework.
\newblock \emph{Proceedings of the 25th ACM SIGKDD International Conference on
  Knowledge Discovery \& Data Mining}.

\bibitem[{Biran et~al.(2011)Biran, Brody, and Elhadad}]{Biran:11}
Or~Biran, Samuel Brody, and No{\'e}mie Elhadad. 2011.
\newblock \href {https://aclanthology.org/P11-2087} {Putting it simply: a
  context-aware approach to lexical simplification}.
\newblock In \emph{Proceedings of the 49th Annual Meeting of the Association
  for Computational Linguistics: Human Language Technologies}, pages 496--501,
  Portland, Oregon, USA. Association for Computational Linguistics.

\bibitem[{De~Belder and Moens(2010)}]{Bel:Moe:10}
Jan De~Belder and Marie-Francine Moens. 2010.
\newblock {Text Simplification for Children}.
\newblock In \emph{Proceedings of the SIGIR Workshop on Accessible Search
  Systems}, pages 19--26.

\bibitem[{De~Belder and Moens(2012)}]{BelderM12}
Jan De~Belder and Marie-Francine Moens. 2012.
\newblock \href {https://doi.org/10.1007/978-3-642-28601-8_36} {{A Dataset for
  the Evaluation of Lexical Simplification}}.
\newblock In \emph{Proceedings of the 13th International Conference on
  Computational Linguistics and Intelligent Text Processing - Volume Part II},
  CICLing'12, page 426–437, Berlin, Heidelberg. Springer-Verlag.

\bibitem[{Devlin et~al.(2019)Devlin, Chang, Lee, and
  Toutanova}]{devlin-etal-2019-bert}
Jacob Devlin, Ming-Wei Chang, Kenton Lee, and Kristina Toutanova. 2019.
\newblock \href {https://doi.org/10.18653/v1/N19-1423} {{BERT}: Pre-training of
  deep bidirectional transformers for language understanding}.
\newblock In \emph{Proceedings of the 2019 Conference of the North {A}merican
  Chapter of the Association for Computational Linguistics: Human Language
  Technologies, Volume 1 (Long and Short Papers)}, pages 4171--4186.

\bibitem[{Glava{\v{s}} and {\v{S}}tajner(2015)}]{glavas-stajner-2015}
Goran Glava{\v{s}} and Sanja {\v{S}}tajner. 2015.
\newblock \href {https://doi.org/10.3115/v1/P15-2011} {{Simplifying Lexical
  Simplification: Do We Need Simplified Corpora?}}
\newblock In \emph{Proceedings of the 53rd Annual Meeting of the Association
  for Computational Linguistics and the 7th International Joint Conference on
  Natural Language Processing (Volume 2: Short Papers)}, pages 63--68, Beijing,
  China. Association for Computational Linguistics.

\bibitem[{Horn et~al.(2014)Horn, Manduca, and
  Kauchak}]{horn-etal-2014-learning}
Colby Horn, Cathryn Manduca, and David Kauchak. 2014.
\newblock \href {https://doi.org/10.3115/v1/P14-2075} {{Learning a Lexical
  Simplifier Using {W}ikipedia}}.
\newblock In \emph{Proceedings of the 52nd Annual Meeting of the Association
  for Computational Linguistics (Volume 2: Short Papers)}, pages 458--463,
  Baltimore, Maryland. Association for Computational Linguistics.

\bibitem[{Lewis et~al.(2019)Lewis, Liu, Goyal, Ghazvininejad, Mohamed, Levy,
  Stoyanov, and Zettlemoyer}]{bart2019}
Mike Lewis, Yinhan Liu, Naman Goyal, Marjan Ghazvininejad, Abdelrahman Mohamed,
  Omer Levy, Veselin Stoyanov, and Luke Zettlemoyer. 2019.
\newblock \href {http://arxiv.org/abs/1910.13461} {{BART:} denoising
  sequence-to-sequence pre-training for natural language generation,
  translation, and comprehension}.
\newblock \emph{CoRR}, abs/1910.13461.

\bibitem[{Loshchilov and Hutter(2019)}]{Loshchilov2019DecoupledWD}
Ilya Loshchilov and Frank Hutter. 2019.
\newblock \href {https://openreview.net/forum?id=Bkg6RiCqY7} {Decoupled weight
  decay regularization}.
\newblock In \emph{International Conference on Learning Representations}.

\bibitem[{Martin et~al.(2020)Martin, de~la Clergerie, Sagot, and
  Bordes}]{martin-etal-2020-controllable}
Louis Martin, {\'E}ric de~la Clergerie, Beno{\^\i}t Sagot, and Antoine Bordes.
  2020.
\newblock \href {https://aclanthology.org/2020.lrec-1.577} {Controllable
  sentence simplification}.
\newblock In \emph{Proceedings of the 12th Language Resources and Evaluation
  Conference}, pages 4689--4698.

\bibitem[{Martin et~al.(2022)Martin, Fan, de~la Clergerie, Bordes, and
  Sagot}]{martin2021muss}
Louis Martin, Angela Fan, {\'E}ric de~la Clergerie, Antoine Bordes, and
  Beno{\^\i}t Sagot. 2022.
\newblock \href {https://aclanthology.org/2022.lrec-1.176} {{MUSS}:
  Multilingual unsupervised sentence simplification by mining paraphrases}.
\newblock In \emph{Proceedings of the Thirteenth Language Resources and
  Evaluation Conference}, pages 1651--1664, Marseille, France. European
  Language Resources Association.

\bibitem[{Paetzold and Specia(2016{\natexlab{a}})}]{Paetzold2016BenchmarkingLS}
Gustavo Paetzold and Lucia Specia. 2016{\natexlab{a}}.
\newblock {Benchmarking Lexical Simplification Systems}.
\newblock In \emph{Proceedings of LREC-2016}.

\bibitem[{Paetzold and Specia(2016{\natexlab{b}})}]{Paetzold_Specia_2016UNS}
Gustavo Paetzold and Lucia Specia. 2016{\natexlab{b}}.
\newblock \href {https://ojs.aaai.org/index.php/AAAI/article/view/9885}
  {{Unsupervised Lexical Simplification for Non-Native Speakers}}.
\newblock \emph{Proceedings of the AAAI Conference on Artificial Intelligence},
  30(1).

\bibitem[{Paetzold and Specia(2017{\natexlab{a}})}]{paetzoldsurvey2017}
Gustavo Paetzold and Lucia Specia. 2017{\natexlab{a}}.
\newblock \href {https://doi.org/10.1613/jair.5526} {{A Survey on Lexical
  Simplification}}.
\newblock \emph{Journal of Artificial Intelligence Research}, 60:549--593.

\bibitem[{Paetzold and
  Specia(2017{\natexlab{b}})}]{paetzold-specia-2017-lexical}
Gustavo Paetzold and Lucia Specia. 2017{\natexlab{b}}.
\newblock \href {https://www.aclweb.org/anthology/E17-2006} {Lexical
  simplification with neural ranking}.
\newblock In \emph{Proceedings of the 15th Conference of the {E}uropean Chapter
  of the Association for Computational Linguistics: Volume 2, Short Papers},
  pages 34--40, Valencia, Spain. Association for Computational Linguistics.

\bibitem[{Qiang et~al.(2020)Qiang, Li, Yi, Yuan, and Wu}]{qiang2020BERTLS}
Jipeng Qiang, Yun Li, Zhu Yi, Yunhao Yuan, and Xindong Wu. 2020.
\newblock Lexical simplification with pretrained encoders.
\newblock \emph{Thirty-Fourth AAAI Conference on Artificial Intelligence},
  pages 8649–--8656.

\bibitem[{Qiang et~al.(2021)Qiang, Li, Zhu, Yuan, Shi, and Wu}]{QiangLSBertACM}
Jipeng Qiang, Yun Li, Yi~Zhu, Yunhao Yuan, Yang Shi, and Xindong Wu. 2021.
\newblock \href {https://doi.org/10.1109/TASLP.2021.3111589} {Lsbert: Lexical
  simplification based on bert}.
\newblock \emph{IEEE/ACM Transactions on Audio, Speech, and Language
  Processing}, 29:3064--3076.

\bibitem[{Raffel et~al.(2020)Raffel, Shazeer, Roberts, Lee, Narang, Matena,
  Zhou, Li, and Liu}]{2020t5}
Colin Raffel, Noam Shazeer, Adam Roberts, Katherine Lee, Sharan Narang, Michael
  Matena, Yanqi Zhou, Wei Li, and Peter~J. Liu. 2020.
\newblock \href {http://jmlr.org/papers/v21/20-074.html} {Exploring the limits
  of transfer learning with a unified text-to-text transformer}.
\newblock \emph{Journal of Machine Learning Research}, 21(140):1--67.

\bibitem[{Saggion(2017)}]{Saggion'2017}
Horacio Saggion. 2017.
\newblock \href {https://doi.org/10.2200/S00700ED1V01Y201602HLT032}
  {\emph{Automatic Text Simplification}}.
\newblock Synthesis Lectures on Human Language Technologies. Morgan {\&}
  Claypool Publishers.

\bibitem[{Shardlow(2014)}]{shardlowsurvey2014}
Matthew Shardlow. 2014.
\newblock \href {https://doi.org/10.14569/SpecialIssue.2014.040109} {{A Survey
  of Automated Text Simplification}}.
\newblock \emph{International Journal of Advanced Computer Science and
  Applications}, 4.

\bibitem[{Sheang and Saggion(2021)}]{sheang-saggion-2021-controllable}
Kim~Cheng Sheang and Horacio Saggion. 2021.
\newblock \href {https://aclanthology.org/2021.inlg-1.38} {Controllable
  sentence simplification with a unified text-to-text transfer transformer}.
\newblock In \emph{Proceedings of the 14th International Conference on Natural
  Language Generation}, pages 341--352, Aberdeen, Scotland, UK. Association for
  Computational Linguistics.

\bibitem[{Wolf et~al.(2020)Wolf, Debut, Sanh, Chaumond, Delangue, Moi, Cistac,
  Rault, Louf, Funtowicz, Davison, Shleifer, von Platen, Ma, Jernite, Plu, Xu,
  Le~Scao, Gugger, Drame, Lhoest, and Rush}]{wolf-etal-2020-transformers}
Thomas Wolf, Lysandre Debut, Victor Sanh, Julien Chaumond, Clement Delangue,
  Anthony Moi, Pierric Cistac, Tim Rault, Remi Louf, Morgan Funtowicz, Joe
  Davison, Sam Shleifer, Patrick von Platen, Clara Ma, Yacine Jernite, Julien
  Plu, Canwen Xu, Teven Le~Scao, Sylvain Gugger, Mariama Drame, Quentin Lhoest,
  and Alexander Rush. 2020.
\newblock \href {https://doi.org/10.18653/v1/2020.emnlp-demos.6} {Transformers:
  State-of-the-art natural language processing}.
\newblock In \emph{Proceedings of the 2020 Conference on Empirical Methods in
  Natural Language Processing: System Demonstrations}, pages 38--45.

\bibitem[{Yatskar et~al.(2010)Yatskar, Pang, Danescu-Niculescu-Mizil, and
  Lee}]{Yatskar10}
Mark Yatskar, Bo~Pang, Cristian Danescu-Niculescu-Mizil, and Lillian Lee. 2010.
\newblock \href {https://aclanthology.org/N10-1056} {For the sake of
  simplicity: Unsupervised extraction of lexical simplifications from
  {W}ikipedia}.
\newblock In \emph{Human Language Technologies: The 2010 Annual Conference of
  the North {A}merican Chapter of the Association for Computational
  Linguistics}, pages 365--368, Los Angeles, California. Association for
  Computational Linguistics.

\bibitem[{Yimam et~al.(2018)Yimam, Biemann, Malmasi, Paetzold, Specia, Stajner,
  Tack, and Zampieri}]{yimametal2018}
Seid~Muhie Yimam, Chris Biemann, Shervin Malmasi, Gustavo~H. Paetzold, Lucia
  Specia, Sanja Stajner, Ana{\"{\i}}s Tack, and Marcos Zampieri. 2018.
\newblock \href {http://arxiv.org/abs/1804.09132} {{A Report on the Complex
  Word Identification Shared Task 2018}}.
\newblock \emph{CoRR}, abs/1804.09132.

\bibitem[{Štajner et~al.(2012)Štajner, Evans, Orasan, and
  Mitkov}]{Stajner&al'2012}
Sanja Štajner, Richard Evans, Constantin Orasan, and Ruslan Mitkov. 2012.
\newblock What can readability measures really tell us about text complexity.
\newblock In \emph{Proceedings of workshop on natural language processing for
  improving textual accessibility}, pages 14--22. Citeseer.

\bibitem[{Štajner et~al.(2022{\natexlab{a}})Štajner, Ferrés, Shardlow,
  North, Zampieri, and Saggion}]{frontiers2022}
Sanja Štajner, Daniel Ferrés, Matthew Shardlow, Kai North, Marcos Zampieri,
  and Horacio Saggion. 2022{\natexlab{a}}.
\newblock \href {https://doi.org/10.3389/frai.2022.991242} {{Lexical
  Simplification Benchmarks for English, Portuguese, and Spanish}}.
\newblock \emph{Frontiers in Artificial Intelligence}, 5.

\bibitem[{Štajner et~al.(2022{\natexlab{b}})Štajner, Sheang, and
  Saggion}]{stajner_sheang_saggion_2022}
Sanja Štajner, Kim~Cheng Sheang, and Horacio Saggion. 2022{\natexlab{b}}.
\newblock \href {https://doi.org/10.1609/aaai.v36i11.21477} {Sentence
  simplification capabilities of transfer-based models}.
\newblock \emph{Proceedings of the AAAI Conference on Artificial Intelligence},
  36(11):12172--12180.

\end{thebibliography}
\bibliographystyle{acl_natbib}

\appendix
 \clearpage
 
\section{Implementation Details} \label{appendix:implementation_details}
Our implementation is based on Huggingface Transformers \cite{wolf-etal-2020-transformers} and Pytorch-lightning\footnote{https://www.pytorchlightning.ai}. We trained the model using T5-Large for 8 epochs. For the optimization, we used AdamW \cite{Loshchilov2019DecoupledWD} optimizer with the learning rate of 1e-5 and adam epsilon of 1e-8. We set the batch size of 8 for both training and testing. For the inference, we used beam search with the size of 15 to get around 10 candidates after filtering out duplicate candidates or the candidates that are the same as the complex word. We trained the model on a machine with an NVidia RTX 3090, Intel core i9 CPU, with 32G of RAM. It took around 2 hours for the whole process: the training and the evaluation on the three datasets.


\end{document}